\documentclass{llncs}
\usepackage{makeidx}  % allows for indexgeneration
\usepackage{llncsdoc}
\usepackage{graphicx}
\usepackage{amssymb}
\usepackage{siunitx}  % SI
\usepackage{textcomp}
\usepackage{adjustbox}
\usepackage{comment}
\usepackage{todonotes}
\usepackage{color}
\usepackage{mathtools}

\usepackage{amsmath}

\usepackage{array}
\newcolumntype{L}[1]{>{\raggedright\let\newline\\\arraybackslash\hspace{0pt}}m{#1}}
\newcolumntype{C}[1]{>{\centering\let\newline\\\arraybackslash\hspace{0pt}}m{#1}}
\newcolumntype{R}[1]{>{\raggedleft\let\newline\\\arraybackslash\hspace{0pt}}m{#1}}

\begin{document}
%%%%%%%%% TITLE %%%%%%%%% 
\title{Assessing the Impact of Blood Pressure on Cardiac Function Using Interpretable Biomarkers and Variational Autoencoders} 

\author{Esther Puyol-Ant\'on* \inst{1} \and 
Bram Ruijsink \thanks{Joint first authors.} \inst{1,2} \and
James R. Clough \inst{1} \and 
Ilkay Oksuz  \inst{1} \and
Daniel Rueckert \inst{3} \and
Reza Razavi  \inst{1,2} \and
Andrew P. King \inst{1}}
\authorrunning{E Puyol-Ant\'on et al.}   % abbreviated author list (for running head)
\institute{School of Biomedical Engineering \& Imaging Sciences, King\textquotesingle s College London, UK. \and St Thomas\textquotesingle{} Hospital NHS Foundation Trust, London, UK. \and Biomedical Image Analysis Group, Imperial College London, UK.}

\maketitle              

\begin{abstract} 
Maintaining good cardiac function for as long as possible is a major concern for healthcare systems worldwide and there is much interest in learning more about the impact of different risk factors on cardiac health. The aim of this study is to analyze the impact of systolic blood pressure (SBP) on cardiac function while preserving the interpretability of the model using known clinical biomarkers in a large cohort of the UK Biobank population. We propose a novel framework that combines deep learning based estimation of interpretable clinical biomarkers from cardiac cine MR data with a variational autoencoder (VAE). The VAE architecture integrates a regression loss in the latent space, which enables the progression of cardiac health with SBP to be learnt. Results on 3,600 subjects from the UK Biobank show that the proposed model allows us to gain important insight into the deterioration of cardiac function with increasing SBP, identify key interpretable factors involved in this process, and lastly exploit the model to understand patterns of positive and adverse adaptation of cardiac function.\\
\keywordname{Cardiac function; variational autoencoder; cardiac risk factors}
\end{abstract}

%%%%%%%%%%%%%%%%%%%%%%%%%%%%%%%%%%%%%%%%%%%%%%%%%%%%%%%%%%%%%%%%%%%%%%%%
% Introduction
%%%%%%%%%%%%%%%%%%%%%%%%%%%%%%%%%%%%%%%%%%%%%%%%%%%%%%%%%%%%%%%%%%%%%%%%
\section{Introduction}
Preventing the development of heart disease in patients with known risk factors, such as hypertension, represents a major challenge for healthcare systems worldwide. Although much is known about how these risk factors influence development of disease, the recent availability of large scale databases such as the UK Biobank represents an excellent opportunity to extend this knowledge. Learning from a large number of highly detailed, multidimensional cardiac magnetic resonance (CMR) datasets can help further understanding of how risk factors impact cardiac function, and tailor medical interventions to individual patients.\\
Traditionally, machine learning techniques have relied on the use of handcrafted features to effectively perform a specific task without using explicit instructions. In some cases, the accuracy of these models was limited by the model being restricted to the use of these features. Recently, deep learning (DL) techniques have demonstrated a significant increase in performance over traditional machine learning methods. DL allows features to be learned from the data themselves, without preselection.
One drawback of DL approaches is the lack of interpretability, as the learned relationships and features are often abstract and opaque to human users. Especially in medicine, interpretability and accountability are vital for two main reasons: (a) they can promote clinician trust in the learned models; and (b) the use of well-established, interpretable biomarkers allows the models to be used to better understand disease processes, translate the results to other populations and to interpret the newly learned information in the light of already existing clinical scientific evidence.  For example, complex full 3D cardiac motion of the heart can be used to outperform current models in survival estimation for patients with pulmonary hypertension \cite{bello2019deep}. While methods like this demonstrate the power of DL, it is difficult for clinicians to understand the features underlying the predictions and to use the model to better understand the disease. \\
A large number of biomarkers can be calculated from CMR. These are well-understood by clinicians and provide comprehensive information about underlying physiological processes. However, estimating them all is labour-intensive.
In this paper, we employ a fully automated DL-based pipeline for estimating a wide range of biomarkers of cardiac function from cine CMR data. Our main contribution is to propose a framework that enables the interpretability of these automatically computed biomarkers to be combined with the power of learned features in DL. This framework is based on the use of a variational autoencoder with a latent space regression loss (R-VAE), in which the input data are the clinical biomarkers. In addition, we use a dummy variable in the regression to differentiate between population groups. We use the proposed method to investigate the impact of systolic blood pressure (SBP: a measure of hypertension) on cardiac function in the healthy population differentiated by gender.\\
\indent \textbf{Related Work:} In the clinical literature, many groups have investigated the relationship between SBP and ventricular structure, function and geometry \cite{Bajpai,Mo}. However, most studies only investigated the influence of SBP on global parameters or only in the left ventricle.  The proposed pipeline enables a more detailed investigation of the impact of SBP on cardiac function, as we demonstrate in Section \ref{sec:results}.
VAEs have previously been used for identification and visualization of features in medical image-based classification tasks \cite{Biffi,Biffi2,Clough}, but the features were still learnt from the data and were not well-established clinical biomarkers as in our work.
DL models have also previously been proposed for regression using interpretable features. For example, Xie \textit{et al.} \cite{xie2018autoencoder} proposed an autoencoder-based deep belief regression network to forecast daily particulate matter concentrations. Similarly, Bose \textit{et al.} \cite{bose2018machine} proposed a stacked autoencoder based regression framework to optimize process control and productivity in intelligent manufacturing. Both techniques combined handcrafted features obtained from the image domain with autoencoders. In the medical field, Xie \textit{et al.} \cite{xie2017deep} have proposed a deep autoencoder model for regression of gene expression profiles from genotype.
Similar to these works we integrate a regression loss into the autoencoder to learn relationships between the latent space and another variable (SBP in our case). Our work is methodologically distinct from \cite{xie2018autoencoder,bose2018machine,xie2017deep} as we employ a VAE, which enables us to sample from the distribution of the latent space and decode the clinical biomarkers for these samples. We also employ a dummy variable in the regression to enable the investigation to be stratified by gender. Our work also represents the first time that a regression-based autoencoder has been applied to investigate the impact of risk factors on cardiac function.
%%%%%%%%%%%%%%%%%%%%%%%%%%%%%%%%%%%%%%%%%%%%%%%%%%%%%%%%%%%%%%%%%%%%%%%%
% Materials and Methods
%%%%%%%%%%%%%%%%%%%%%%%%%%%%%%%%%%%%%%%%%%%%%%%%%%%%%%%%%%%%%%%%%%%%%%%%
\section{Materials}
\label{sec:materials}
We evaluate our approach on subjects selected from the UK Biobank data set, which contains multiple imaging and non-imaging information from more than half a million 40-69 year-olds. We included only participants with CMR imaging data. From this group, we excluded participants with a history of cardiovascular disease, respiratory disease, renal disease, cancer, rheumatological disease, symptoms of chest pain or dyspnoea. For each subject, the following cine CMR acquisitions were used; a short-axis (SA) stack covering the full heart, and two orthogonal long-axis (LA) acquisitions (2-chamber (2Ch) and 4-chamber (4Ch) views). For each image slice 50 temporal frames were available covering a full cardiac cycle (temporal resolution $\approx$14-24 ms/frame). All CMR imaging was carried out on a 1.5 Tesla scanner (Siemens Healthcare, Erlangen, Germany). Details of the image acquisition protocol can be found in \cite{petersen2016uk}. Blood pressure was  measured using the HEM-70151T digital blood pressure monitor (Omron, Hoofddorp, The Netherlands) \cite{Chan}
%%%%%%%%%%%%%%%%%%%%%%%%%%%%%%%%%%%%%%%%%%%%%%%%%%%%%%%%%%%%%%%%%%%%%%%%
% Methods
%%%%%%%%%%%%%%%%%%%%%%%%%%%%%%%%%%%%%%%%%%%%%%%%%%%%%%%%%%%%%%%%%%%%%%%%
\section{Methods}
\label{sec:methods}
\begin{figure}[ht]
\centering
\includegraphics[width=\textwidth]{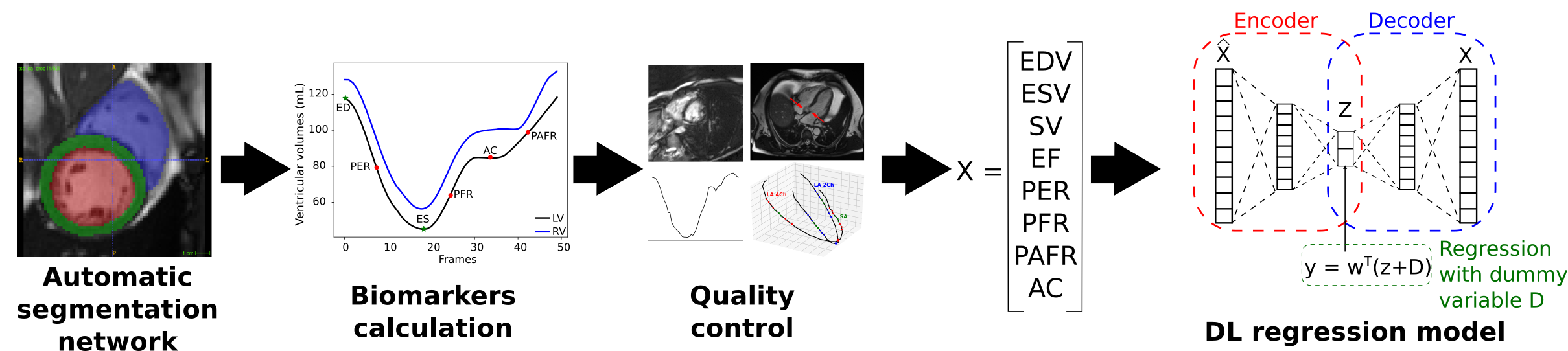}
\caption{Overview of the proposed framework for a VAE regression model based on automatically estimated clinical biomarkers.}
\label{fig:pipeline}
\end{figure}

In the following sections we first describe the automated estimation of biomarkers of cardiac function from CMR images, and secondly present the R-VAE network used to learn the relationship between cardiac function and SBP. Figure \ref{fig:pipeline} summarizes these steps and how they interrelate.

\subsection{Biomarker Estimation From CMR}
\label{subsec:clinical_descriptors}
The procedure used to extract the clinical biomarkers of cardiac function is based on the work published in \cite{jacc2018EB} and is briefly summarized below.\\
\indent \textbf{Automatic segmentation network:} We first used a fully-convolutional network with a 17 convolutional layer VGG-like architecture for automatic segmentation of the left ventricle (LV) blood pool, LV myocardium and right ventricle (RV) blood pool from SA and LA slices in all frames through the cardiac cycle \cite{bai2017semi,sinclair2017fully}. After this, all segmentations were aligned to correct for breath-hold induced motion artefacts using the iterative registration algorithm proposed in \cite{sinclair2017fully}.\\
\indent \textbf{Biomarker calculation:} LV and RV blood volume curves were calculated from the obtained segmentations. From these curves, end-diastolic volume (EDV), end-systolic volume (ESV), stroke volume (SV), ejection fraction (EF), LVED mass (LVEDM), peak ejection rate (PER), peak early filling rate (PEFR), atrial contribution (AC) and peak atrial filling rate (PAFR) were obtained. Cardiac volumes were indexed to body surface area (BSA) using the Dubois and Dubois formula \cite{dubois}. The complete list of biomarkers calculated was: iLVEDV (indexed LVEDV), iLVSV, LVEDM, LVPER, LVPFR, LVPAFR, LVAC, iRVEDV, RVPER, RVPFR, RVPAFR and RVAC.
\\
\indent \textbf{Quality control:} Similar to \cite{jacc2018EB}, two quality control (QC) methods were implemented to automatically reject subjects with insufficient image quality or incorrect segmentations, ensuring the robustness of the estimated biomarkers. The first QC step (QC1) used trained DL models to reject any image with poor quality or incorrect planning, and the second QC step (QC2) detected incorrect segmentation results using an SVM model that identified physiologically unrealistic or unusual volume curves.
\subsection{Deep Learning Regression Model}
\label{subsec:dl_regression}
To combine the interpretability of handcrafted features with the power of DL we propose to use a VAE featuring a regression loss in the latent space to simultaneously learn efficient representations of cardiac function and map their change with regard to differences in SBP. As a regression model we used the multivariable regression modelling commonly used in epidemiological studies \cite{Thompson}, where the effect of different independent variables are included as confounders on the regression model. As a result, the VAE linearizes the relationship between different clinical biomarkers and the variable of interest (SBP in this case), and these features are incorporated in a standard regression model.\\
\indent \textbf{Variational autoencoder:}
The encoding part of a VAE allows a number of features, $\mathbf{x}$, to be mapped into a lower dimensional representation (the \emph{latent space}), whilst the decoder maps this representation back to the original higher dimensional space. The proposed VAE has $N=13$ input units representing the clinical biomarkers, two hidden layers with 8 and 4 hidden units respectively and a latent space of dimensionality 2. To avoid over-fitting we apply dropout with probability 0.3 after each hidden layer during training.\\
\indent \textbf{Linear regression  with indicator (dummy) variable:} We use a linear regression model in the latent space to estimate SBP, that incorporates a dummy variable (encoded as 0 or 1) to differentiate between population groups. In our experiments we used gender as a dummy variable, but the method is general and could be used to investigate a wide range of other factors. Mathematically, the linear regression model can be formulated as follows: $y = w^T(z+D) + \epsilon$, where $z$ are the latent space activations, $D$ the dummy variable, $y$ the ground truth label (i.e. SBP) and $w$ the regression coefficients. The regression loss is the mean squared error: $L_{\mathrm{regression}} = \displaystyle \frac{1}{N}\sum_{t=1}^{N} (y_i-w_i^T(z_i+D_i))^2$.\\ 
\indent \textbf{Joint learning for regression:} We denote the input data by $\mathbf{x}=[x_1, x_2, ... x_N]$
(i.e. a vector of $N=13$ clinical biomarkers) and its corresponding latent space representation as $\mathbf{z}= [z_1, z_2]$.The decoded clinical biomarkers are denoted by $\mathbf{\hat{x}}= [\hat{x}_1, \hat{x}_2, ... \hat{x}_N]$, and the predicted label by $\hat{y} = \text{Regressor}(\mathbf{z}).$ We combine the VAE loss and the regression loss by minimising the following joint loss function:
\begin{equation}
    L_{\mathrm{R-VAE}}
    = L_{\mathrm{recon}} +  \alpha L_{\mathrm{KL}} + \beta L_{\mathrm{regression}}
\end{equation}
where $\alpha$ and $\beta$ control the weights of the components of the loss function, $L_{\mathrm{KL}}$ is the Kullback-Leibler divergence between the learnt latent distribution and a unit Gaussian, and $L_{\mathrm{regression}}$ is the Huber loss for the regression task. We first train the model only using the VAE loss, i.e. $\beta=0$, and secondly train both the VAE and the regression together using $\beta=2$. We set $\alpha=0.3$ throughout.
%%%%%%%%%%%%%%%%%%%%%%%%%%%%%%%%%%%%%%%%%%%%%%%%%%%%%%%%%%%%%%%%%%%%%%%%
% Experiments and Results
%%%%%%%%%%%%%%%%%%%%%%%%%%%%%%%%%%%%%%%%%%%%%%%%%%%%%%%%%%%%%%%%%%%%%%%%
\section{Experiments and Results}
\label{sec:results} 
Using our selection criteria, 3,781 subjects were included in our experiments. Of these, 54 subjects were rejected during QC1, and a further 127 during QC2. The remaining 3,600 cases were used to build the models. Of these cases, 1,321 had normotension (SBP $<$120 mmHg), 1,697 hypertension (SBP $>$140 mmHg) and 582 cases had prehypertension (a SBP between 120 and 140 mmHg).\\
\indent \textbf{Experiment 1 - Comparative evaluation on regression task:}\\
We compared our proposed R-VAE model with two state-of-the-art techniques for multivariate regression: (1) Lasso regression, a linear regression model with a $l1$ regularizer \cite{tibshirani1996regression}; and (2) Random Forest regression \cite{lin2006random}, an ensemble method that has shown excellent performance in complex regression and classification tasks. For all methods, we used a five-fold cross validation to obtain the optimal model. We split the dataset into training, validation and test (60/20/20), and optimized the hyperparameters using a grid search strategy.
We calculated the root-mean-square deviation (RMSD), the normalized root-mean-square deviation (nRMSD), and the coefficient of determination ($R^2$) between the ground truth SBP and the predicted SBP. 
Table \ref{table:1} shows these results. It can be seen that all methods performed similarly with regard to regression. However, note that the next two experiments are only made possible by our use of the R-VAE architecture and would not be possible with the other regression techniques.\\

\begin{table} [!h]
\centering
\caption{Comparison of R-VAE, Lasso regression and Random Forest regression.}
\begin{tabular}{L{3cm} C{2.7cm} C{2.7cm} C{2.7cm}}
\hline
Methods   & RMSD (mmHg) & nRMSD (mmHg) & $R^2 (\%)$   \\ \hline 
Lasso \cite{tibshirani1996regression} & 13.2  & 0.11 & 0.35 \\
Random Forest  \cite{lin2006random} & 12.98  & 0.12 & 0.33 \\
R-VAE & 11.36 & 0.10 & 0.69 \\
\hline
\end{tabular}
\label{table:1}
\end{table}

\begin{figure}[ht]
\centering
\includegraphics[width=\textwidth]{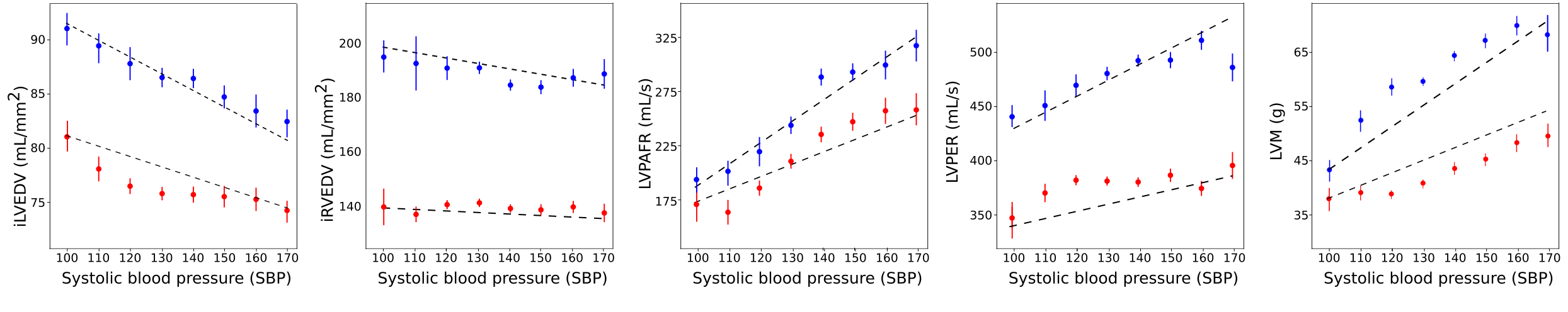}
\caption{SBP-related changes in iLVEDV, iRVEDV, LVPAFR and LVPER. Red represents females and blue represents males. Bars represent standard deviations. Black dotted lines represent the linear tendency curves between the cardiac biomarkers and ground-truth SBP data.}
\label{fig:vols}
\end{figure}

\indent \textbf{Experiment 2 - Investigating the effect of SBP on cardiac function:}\\
We applied our R-VAE network to investigate how cardiac function changes with increasing SBP in healthy individuals, stratified by gender. We applied our R-VAE network to investigate how cardiac function changes with increasing SBP in healthy individuals, stratified by gender. We sampled the latent space of the model along the regression line for different SBP values (between 100 to 170 mmHg in steps of 10 mmHg). At each step, we took 20 random samples from a normal distribution in a perpendicular direction to the regression line and used the R-VAE to decode the clinical biomarkers. Figure \ref{fig:vols} shows the means and standard deviations for a selection of the decoded  biomarkers stratified by gender. The results show that iLVEDV (indexed LVEDV) decreases with increasing SBP, while iRVEDV (indexed RVEDV) remains constant. LVPAFR and LVPER increase with increasing SBP. For both iLVEDV and LVPER, the change seems to be larger in males compared to females. Overall, the observed changes in the models' predictions suggest that parameters associated with diastolic function of the LV are mostly affected by SBP, while the RV was less affected. These results suggest that stiffening of the LV myocardium could be an important disease process in deterioration of cardiac function in the light of increased SBP.

\begin{figure}[ht]
\centering
\includegraphics[width=\textwidth]{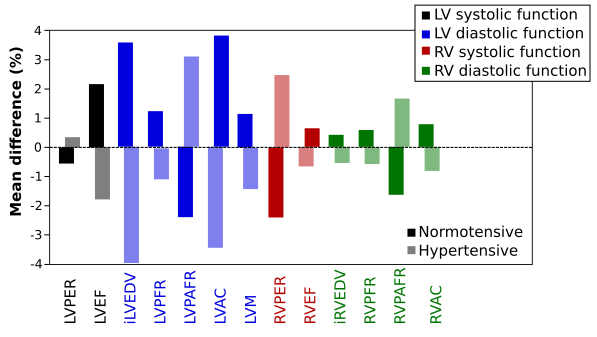}
\caption{Mean change (percentage) of each biomarker in prehypertension cases that were classified by the regression model as normotensive (dark) and hypertensive (light) with respect to values predicted by the model using the actual observed SBP. Values further away from zero mean a larger impact of these biomarkers.}
\label{fig:pred}
\end{figure}

\indent  \textbf{Experiment 3 - Identifying abnormal response:}\\
In the normal population, some individuals with prehypertension might be predisposed to increased risk of cardiac disease, while others are not. We used our R-VAE model to identify subjects from the prehypertension group in whom predicted SBP was lower (i.e. predicted normotension) or higher (i.e. predicted hypertension) based on the latent space regression, assuming that being wrongly classified as normotensive or hypertensive  identifies individuals with low versus increased risk of developing cardiac disease. Subsequently, we decoded the cardiac biomarkers for these cases using latent features at the regression line of their true SBP as well as using latent features at their predicted SBP. We calculated the percentage difference for each biomarker for the cases of under- and over-prediction respectively and investigated which factors contributed most to the lower or higher prediction in these subjects.
Figure \ref{fig:pred} shows the mean difference of selected biomarkers that lead to classification as normotensive (dark) and hypertensive (light). Biomarkers related to LV diastolic function (blue bars)  showed the largest changes with regard to the under- and overprediction. These results show again that diastolic function of the LV was a major contributor to the model predictions of SBP. Moreover, they suggest that biomarkers related to LV diastolic function might be effective when trying to stratify cardiac risk, in particular in subjects with `prehypertension'.  

%%%%%%%%%%%%%%%%%%%%%%%%%%%%%%%%%%%%%%%%%%%%%%%%%%%%%%%%%%%%%%%%%%%%%%%%
% Discussion
%%%%%%%%%%%%%%%%%%%%%%%%%%%%%%%%%%%%%%%%%%%%%%%%%%%%%%%%%%%%%%%%%%%%%%

\section{Discussion}
\label{sec:discussion}
In this paper, we have presented an automated DL method for analysing cardiac function and predicting cardiac risk profiles from CMR. Our framework encompasses all steps from CMR segmentation through quality control to modelling of the impact of SBP, a known cardiovascular risk factor, on cardiac function.\\
%using a technique that allows direct interpretability of the results within the clinical context. 
Instead of inputting CMR images directly into the R-VAE, we chose to first automatically estimate clinical biomarkers from the images. Combined, these biomarkers give a comprehensive description of cardiac function and are well-understood  by clinicians. While some of the information of the high dimensional image data is inevitably lost by this approach, it allows the model to be interpretable by clinicians directly.\\  
%The performance of our method was similar to Random Forest, the state-of-the-art for regression taks. However, our approach provides better interpretability compares to a Random Forest. 
The combination of the VAE with the regression loss allowed us to decode the clinical biomarkers from the latent space, while also providing a mapping of these biomarkers to another variable, SBP.
As we show in Experiment 2, this design enabled us to get a clear description of the changes in cardiac function that occur with increasing SBP. Using the trained model, we showed that increases in SBP are mainly linked to changes in diastolic LV function. This suggests that SBP results in slowly progressive changes in the myocardium that increase ventricular stiffening. As shown in Experiment 3, the model also allowed us to identify key factors separating high and low risk subjects. Again, due to the interpretability of the framework, this allowed us to identify biomarkers that could be further investigated for their utility in screening patients in clinical practice.
SBP is not the only factor influencing cardiac function and that explains the relatively low $R^2$ of the regression models. In this paper, we used SBP as an example to illustrate the potential power of our proposed method. We aim to further extend our model in the following ways: we plan to include image and segmentation data directly into the model, in combination with the clinical biomarkers to maintain interpretability; we also plan to extend the model to investigate more risk factors. In conclusion, this work represents a novel use of DL which has produced an important contribution to furthering our understanding of the influences on cardiac function.

\section*{Acknowledgements}
This work was supported by the EPSRC (grants EP/R005516/1 and EP/P001009/1) and the Wellcome EPSRC Centre for Medical Engineering at the School of Biomedical Engineering and Imaging Sciences, King’s College London (WT 203148/Z/16/Z). This research has been conducted using the UK Biobank Resource under Application Number 17806.

%%%%%%%%%%%%%%%%%%%%%%%%%%%%%%%%%%%%%%%%%%%%%%%%%%%%%%%%%%%%%%%%%%%%%%%%
% References
%%%%%%%%%%%%%%%%%%%%%%%%%%%%%%%%%%%%%%%%%%%%%%%%%%%%%%%%%%%%%%%%%%%%%%
\bibliographystyle{ieeetr}
\bibliography{Biblio}

\end{document}